\documentclass{article}

\usepackage{amssymb}
\usepackage{amsmath}
\DeclareMathOperator*{\argmax}{arg\!max}
\usepackage{authblk}
\usepackage{graphicx}
\usepackage{fancyhdr}

\usepackage{ijcai16}

\usepackage{times}

\title{Classifying Options for Deep Reinforcement Learning}
\author[1]{\textbf{Kai Arulkumaran}\vspace{-0.5ex}}
\author[2]{\textbf{Nat Dilokthanakul}\vspace{-0.5ex}}
\author[2]{\textbf{Murray Shanahan}\vspace{-0.5ex}}
\author[1]{\textbf{Anil Anthony Bharath}\vspace{-0.5ex}}
\affil[1 ]{Department of Bioengineering}
\affil[2 ]{Department of Computing}
\affil[ ]{Imperial College London, London SW7 2BP, UK}
\affil[ ]{\{kailash.arulkumaran13,n.dilokthanakul14,m.shanahan,a.bharath\}@imperial.ac.uk}

\begin{document}

\maketitle

\begin{abstract}
In this paper we combine one method for hierarchical reinforcement
learning---the options framework---with deep Q-networks (DQNs) through
the use of different ``option heads'' on the policy network, and a
supervisory network for choosing between the different options. We
utilise our setup to investigate the effects of architectural
constraints in subtasks with positive and negative transfer, across a
range of network capacities. We empirically show that our augmented DQN
has lower sample complexity when simultaneously learning subtasks with
negative transfer, without degrading performance when learning subtasks
with positive transfer.
\end{abstract}

\section{Introduction}\label{introduction}

Recent advances in reinforcement learning have focused on using deep
neural networks to represent the state-action value function
(Q-function) \cite{mnih2015human} or a policy function
\cite{levine2015end}. The successes of such methods are largely
attributed to the representational power of deep networks. It is
tempting to create an ultimate end-to-end solution to general problems
with these kinds of powerful models. However, these general methods
require large amount of samples in order to learn effective policies.

In order to reduce sample complexity, one can use domain knowledge to
construct an algorithm that is biased towards promising solutions. The
deep Q-network (DQN) \cite{mnih2015human} is an end-to-end
reinforcement learning algorithm that has achieved great success on a
variety of video games \cite{bellemare2013arcade}. In this work, we
explore the possibility of imposing some structural priors onto the DQN
as a way of adding domain knowledge, whilst trying to limit the
reduction in generality of the DQN.

In classical reinforcement learning literature, there is a large amount
of research focusing on temporal abstraction of problems, which is known
under the umbrella of hierarchical reinforcement learning. The options
framework \cite{sutton1999between} augments the set of admissible
actions with temporally-extended actions that are called options. In
this context, an option is a closed-loop controller that can execute
either \emph{primitive} actions or other options according to its
policy. The ability to use options gives the agent many advantages, such
as temporally-extended exploration, allowing planning across a range of
time scales and the ability to transfer knowledge to different tasks.

Another approach, MAXQ \cite{dietterich2000hierarchical}, decomposes a
task into a hierarchy of subtasks. This decomposition allows MAXQ agents
to improve performance towards achieving the main objective by
recursively solving smaller problems first. Both the options and MAXQ
frameworks are constructed so that prior domain knowledge can be
implemented naturally. Both approaches can be viewed as constructing a
main policy from smaller sub-policies with implemented prior knowledge
on either the structural constraints or the behaviour of the
sub-policies themselves.

We consider the class of problems where the task can be broken down into
reward-independent subtasks by a human expert. The task decomposition is
done such that the subtasks share the same state space, but can be
explicitly partitioned (for different options). If the action space can
similarly be partitioned, then this knowledge can also be incorporated
to bound the actions available to each option policy. With this domain
knowledge we decompose the DQN into a composition of smaller
representations, derived from prior work on hierarchical reinforcement
learning. Although there is evidence that the DQN can implicitly learn
options \cite{zahavy2016graying}, we investigate whether there is any
benefit to constructing options explicitly.

\subsection{Related Work}\label{related-work}

Our ``option heads'' are inspired by Osband \emph{et al.}
\shortcite{osband2016deep}. Their policy network is similar to ours,
but without the additional supervisory network. Their network's
``bootstrap heads'' are used with different motivations. They train each
of the heads, with different initialisations, on the same task. This
allows the network to represent a distribution over Q-functions.
Exploration is done ``deeply'' by sampling one head and using its policy
for the whole episode, while the experiences are shared across the
heads. Our motivation for using option heads, however, is for allowing
the use of temporally abstracted actions akin to the options framework,
or more concretely, as a way to decompose the policy into a combination
of simpler sub-policies. This motivates the use of a supervisory
network, which is discussed in Subsection \ref{supervisory-network}.

Although we focus on the options framework and the notion of
\emph{subtasks}, an alternative view is that of multitask learning. In
the context of deep reinforcement learning, recent work has generalised
\emph{distillation} \cite{hinton2015distilling}, applied to
classification, in order to train DQNs on several Atari games in
parallel \cite{rusu2015policy,parisotto2015actor}. Distillation uses
trained teacher networks to provide extra training signals for the
student network, where originally the technique was used to
\emph{distill} the knowledge from a large teacher network into a smaller
student network as a form of model compression. For multitask learning,
Parisotto \emph{et al.} \shortcite{parisotto2015actor} were able to
keep the student DQN architecture the same as that of the teacher DQNs,
whilst Rusu \emph{et al.} \shortcite{rusu2015policy} created a larger
network with an additional fully connected layer. The latter explicitly
separated the top of the network into different ``controllers'' per
game, and called the architecture a Multi-DQN, and the architecture in
combination with their policy distillation algorithm Multi-Dist. Both
studies showed that the use of teacher networks could enable learning
effective policies on several Atari games in parallel---8 on a standard
DQN \cite{parisotto2015actor} and 10 on a Multi-DQN
\cite{rusu2015policy}. The same architectures were unable to perform
well across all games without teacher networks.

We note that the bootstrapped DQN and the multi-DQN have similar
structures: several ``heads'' either directly or indirectly above shared
convolutional layers. One of the baselines for evaluating the
actor-mimic framework \cite{parisotto2015actor} is the Multitask
Convolutional DQN (MCDQN), which has the same architecture as the
bootstrapped DQN. Although working from the same architecture, our goals
are different due to incorporating the DQN into the options framework.
The first major difference in our work is that we use a supervisory
network, which allows us to infer which subtask should be attempted at
each time step during evaluation. Conversely, in a multitask setting,
different tasks are typically clearly separated. Secondly, our method
does not rely on teacher networks, instead focusing on a DQN whose
augmented training signal is based only on the knowledge of the current
subtask. With respect to the latter point, we focus our analysis on
controlling the capacity of the networks, as opposed to scaling
parameters linearly with the number of tasks
\cite{rusu2015policy,parisotto2015actor}.

Another notable success in subtask learning with multiple independent
sources of reward are universal value function approximators (UVFAs)
\cite{schaul2015universal}. UVFAs allow the generalisation of value
functions across different goals, which helps the agent accomplish tasks
that it has never seen before. The focus of UVFAs is in generalising
between similar subtasks by sharing the representation between the
different tasks. This has recently been expanded upon in the
hierarchical-DQN \cite{kulkarni2016hierarchical}; however, these
goal-based approaches have been demonstrated in domains where the
different goals are highly related. From a function approximation
perspective, goals should share a lot of structure with the raw states.
In contrast, our approach focuses on separating out distinct subtasks,
where partial independence between subpolicies can be enforced through
structural constraints. In particular, we expect that separate
Q-functions are less prone to negative transfer between subtasks.

\section{Background}\label{background}

Consider a reinforcement learning problem, where we want to find an
agent's policy \(\pi\) which maximises the expected discounted reward,
\(\mathbb{E}[R] = \mathbb{E}\left[\sum_t \gamma ^t r_t\right]\). The
discount parameter, \(\gamma \in [0, 1]\), controls the importance of
immediate rewards relative to more distant rewards in the future. The
reward \(r_t\) is a scalar value emitted from a state
\(s_t \in \mathcal{S}\). The policy selects and performs an action
\(a_t \in \mathcal{A}\) in response to the state \(s_t\), which then
transitions to \(s_{t+1}\). The transition of states is modelled as a
Markov decision process (MDP) where each state is a sufficient statistic
of the entire history, so that the transition at time \(t\) need only
depend on \(s_{t-1}\) and the action \(a_{t-1}\). See
\cite{sutton1998reinforcement} for a full introduction.

The Q-learning algorithm \cite{watkins1989learning} solves the
reinforcement learning problem by approximating the optimal state-action
value or Q-function, \(Q^*(s_t,a_t)\), which is defined as the expected
discounted reward starting from state \(s_t\) and taking initial action
\(a_t\), and henceforth following the optimal policy \(\pi^*\):
\[Q^{*}(s_t,a_t) = \mathbb{E}[R_t | s_t, a_t, \pi^*]\] \noindent The
Q-function must satisfy the Bellman equation,
\[Q^{*}(s_t,a_t) = \mathbb{E}_{s_{t+1}}[r_t + \gamma \max_{a_{t+1}}{Q^*(s_{t+1},a_{t+1})}].\]
\noindent We can approximate the Q-function with a function approximator
\(Q(s,a;\theta)\), with parameters \(\theta\). Learning is done by
adjusting the parameters in such a way to reduce the inconsistency
between the left and the right hand sides of the Bellman equation. The
optimal policy can be derived by simply choosing the action that
maximises \(Q^*(s,a)\) at each time step.

\subsection{Deep Q-networks}\label{deep-q-networks}

The DQN \cite{mnih2015human} is a convolutional neural network that
represents the Q-function \(Q(s,a;\theta)\) with parameters \(\theta\).
The (online) network is trained by minimising a sequence of loss
functions at iteration \(i\):

\begin{equation}
\begin{aligned}
L_i(\theta_i) = \mathbb{E}_{s,a,r,s'}[(y_i - Q(s,a;\theta_i))^2] \notag \\
\end{aligned}
\end{equation}\begin{equation}
\begin{aligned}
y_i = r + \gamma \max_{a'}{Q(s',a';\theta^-)}
\label{eq:DQNtarget}
\end{aligned}
\end{equation}

The parameters \(\theta^-\) are associated with a separate target
network, which is updated with \(\theta_i\) every \(\tau\) steps. The
target network increases the stability of the learning. The parameters
\(\theta_i\) are updated with mini-batch stochastic gradient descent
following the gradient of the loss function.

Another key to the successful training of DQNs is the use of experience
replay \cite{lin1992self}. Updating the parameters, \(\theta_i\), with
stochastic gradient descent on the squared loss function implies an
i.i.d. assumption which is not valid in an online reinforcement learning
problem. Experience replay stores samples of past transitions in a pool.
While training, samples are drawn uniformly from this pool. This helps
break the temporal correlation between samples and also allows updates
to reuse samples several times.

\subsection{Double Deep Q-networks}\label{double-deep-q-networks}

We follow the learning algorithm by van Hasselt \textit{et al.}
\shortcite{van2015deep} to lower the overestimation of Q-values in the
update rule. This modifies the original target, Equation
\ref{eq:DQNtarget}, to the following,
\[y_i = r + \gamma Q(s', \argmax_{a'}{Q(s',a';\theta_i)}; \theta^-).\]

\subsection{The Options Framework}\label{the-options-framework}

The orginal definition of options \cite{sutton1999between} consists of
three components: a policy, \(\pi\), a termination condition, \(\beta\),
and an initiation set, \(I\). We illustrate the role of these components
by following the interpretation by Daniel \textit{et al.}
\shortcite{daniel2012hierarchical}.

Consider a stochastic policy, \(\pi(a_t|s_t)\), or a distribution over
actions, \(a_t\), given state, \(s_t\), at time step \(t\). We add an
auxiliary variable, \(o_t \in \mathcal{O}\), such that \(a_t\) is
dependent on \(o_t\), and, \(o_t\) is dependent on \(s_{t}\) and
\(o_{t-1}\). This variable \(o_t\) controls the selection of action
\(a_t\) through their conditional dependence, \(\pi(a_t | s_t, o_t)\),
and can be interpreted as the policy of a Markov option \(o_t\). The
termination condition, \(\beta\), can be thought of as a specific
constraint of the conditional form, imposed on the transition of the
option as follows:

\begin{equation}
\begin{aligned}
\pi(o_t| o_{t-1}, s_t) &\propto \beta(s_t, o_{t-1}) \pi(o_t|s_t) \\ &+ \delta_{o_t,o_{t-1}}(1 - \beta(s_t, o_{t-1})), \notag
\end{aligned}
\end{equation}

\noindent where \(\delta_{o_t,o_{t-1}}\) is 1 when \(o_t = o_{t-1}\) and
0 otherwise. The initiation set specifies the domain of \(s_t\)
available to \(\pi(o_t|s_t)\).

We consider a fully observable MDP where \(s_t\) is assumed to be a
sufficient statistic for all the history before \(t\), including
\(o_{t-1}\). Therefore, we can model \(o_t\) to be conditionally
indepedent of \(o_{t-1}\) given \(s_t\). We define our ``supervisory
policy'' as \(\pi(o_t| s_t)\). Both the termination condition and the
initiation set are absorbed into the supervisory policy. The full policy
is then decomposed into:
\[\pi(a_t| s_t) = \sum_{o_t} \pi(o_t|s_t) \pi(a_t | o_t, s_t).\]
\noindent This form of policy can be seen as a combination of option
policies, weighted by the supervisory policy. We will show in the next
section how we decompose the DQN into separate option policies,
alongside a supervisory policy.

\section{Deep Q-networks with Option
Heads}\label{deep-q-networks-with-option-heads}

\begin{figure}
\centering
\includegraphics[width=0.32000\textwidth]{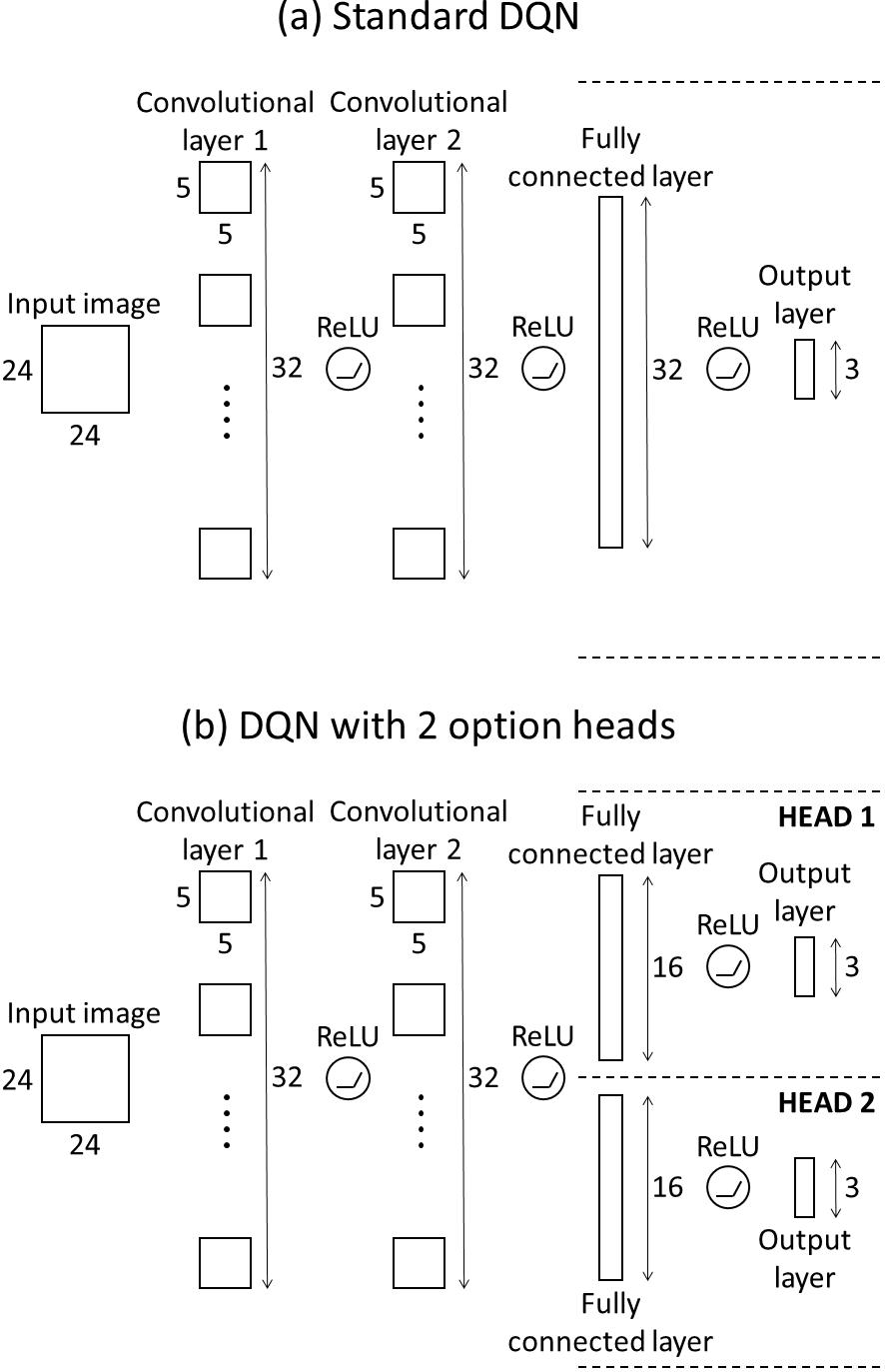}
\caption{Comparison of DQN architectures. a) Standard DQN. b) DQN with 2
option heads.\label{architecture}}
\end{figure}

We augment the standard DQN with several distinct sets of outputs;
concretely we use the same architecture as the bootstrapped DQN
\cite{osband2016deep} or the MCDQN \cite{parisotto2015actor}. As with
the MCDQN, we use domain knowledge to choose the number of heads a
priori, and use this same knowledge to train each option head
separately. A comparison between a standard DQN and a DQN with option
heads that we use is pictured in Figure \ref{architecture}. As noted in
\cite{rusu2015policy}, even this augmentation can fail in a multitask
setting with different policies interfering at the lower levels of the
network, which highlights the need for further study.

Along with other work on the DQN, we assume that the convolutional
layers learn a general representation of the state space, whilst the
fully connected layers at the top of the network encode most of the
actual policy. In the multitask Atari setting, the set of games are so
different that the states are unlikely to have many features in common
\cite{rusu2015policy}, but we would assume that in a hierarchical
reinforcement learning setting with subtasks this problem does not
occur.

In addition to our augmented \emph{policy network},
\(Q_o(s,a) : \mathcal{O} \times \mathcal{S} \times \mathcal{A} \rightarrow \mathbb{R}\),
where \(o\) indexes over option heads, we also introduce a
\emph{supervisory network},
\(O(s) : \mathcal{S} \rightarrow \mathcal{O}\), which learns a mapping
from a state to an option; this allows each option head to focus on a
subset of the state space. With our networks, our full policy can be
written as a deterministic mapping,
\(\pi(s) : \mathcal{S} \rightarrow \mathcal{A}\),
\[\pi(s) = \argmax_a Q_{O(s)}(s,a)\]

\subsection{Option Heads}\label{option-heads}

The option heads consist of fully connected layers which branch out from
the topmost shared convolution layer. The final layer of each head
outputs the Q-value for each discrete action available, and hence can be
limited using domain knowledge of the task at hand and the desired
options. While training, an oracle is used to choose which option head
should be evaluated at each time step \(t\). The action \(a_t\) is
picked with the \(\epsilon\)-greedy strategy on the \(o_t\) head, where
\(\epsilon\) is shared between all heads. The experience samples are
tuples of \((s_t,a_t,s_{t+1},r_t)\), and are stored in separate
experience replay buffers for each head. During evaluation the oracle is
replaced with the decisions of the supervisory network.

\subsection{Supervisory Network}\label{supervisory-network}

The supervisory network is an arbitrary neural network classifier which
represents the supervisory policy. The input layer receives the entire
state. The hidden layers can be constructed using domain knowledge,
e.g.~convolutional layers for visual domains. The output layer is a
softmax layer which outputs the distribution over options, \(o_t\),
given the state, \(s_t\), and can be trained with the standard
cross-entropy loss function. During training the targets are given from
an oracle.

\section{Experiments}\label{experiments}

\begin{figure}
\centering
\includegraphics[width=0.32000\textwidth]{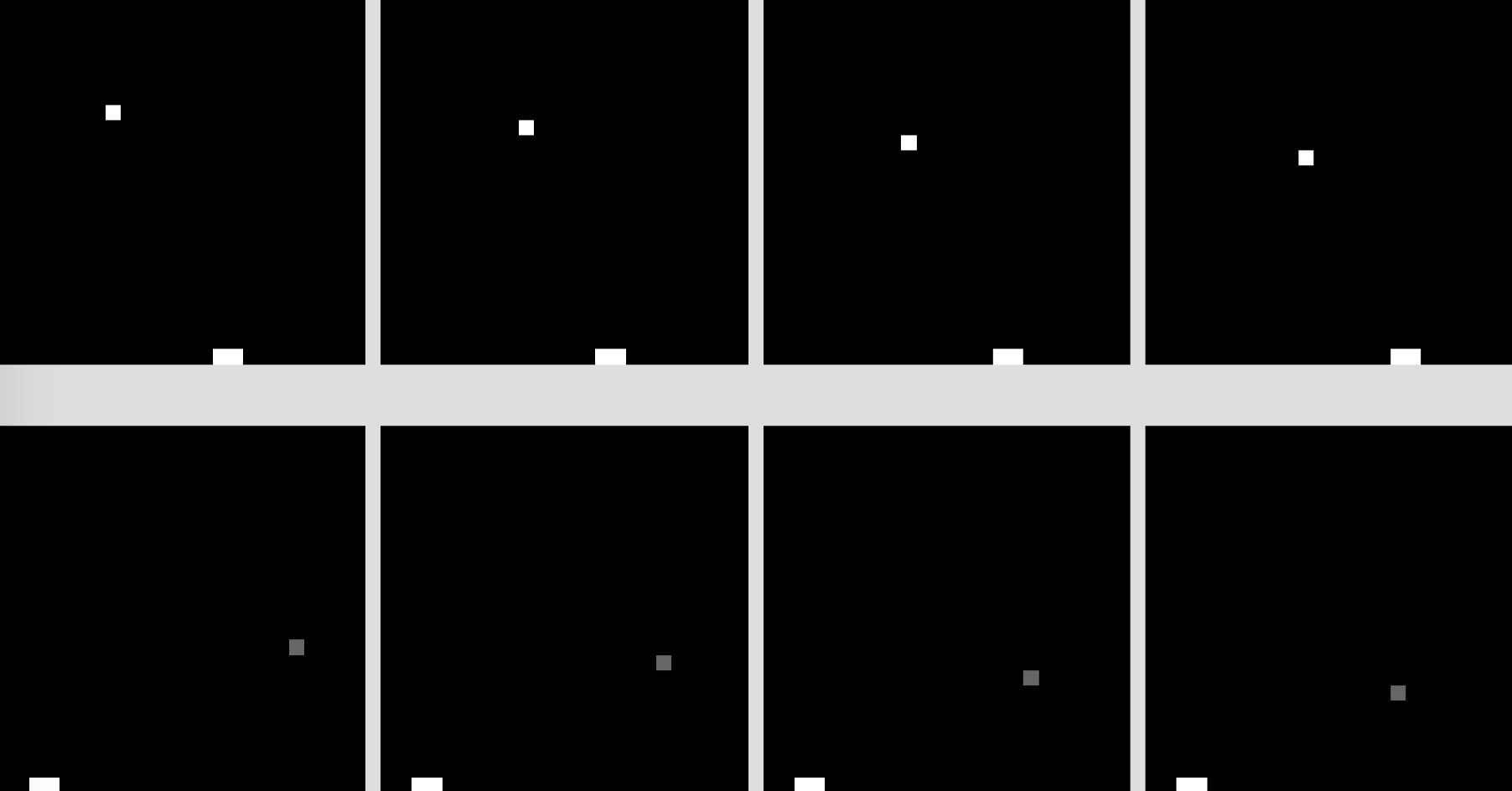}
\caption{8 frames of Catch. The first row shows the white ball subtask,
and the second row shows the grey ball subtask.\label{catch}}
\end{figure}

For our experiments we reimplemented the game of ``Catch''
\cite{mnih2014recurrent}, where the task is to control a paddle at the
bottom of the screen to catch falling balls (see Figure \ref{catch}).
The input is a greyscale 24x24 pixel grid, and the action space consists
of 3 discrete actions: move left, move right and a no-op. As in
\cite{mnih2015human}, the DQN receives a stack of the current plus
previous 3 frames. During each episode a 1 pixel ball falls randomly
from the top of the screen, and the agent's 2-pixel-wide paddle must
move horizontally to catch it. In the original a reward of +1 is given
for catching a white ball; we add an additional grey ball to introduce
subtasks into the environment. This simple environment allows us to
meaningfully evaluate the effects of the architecture on subtasks with
\emph{positive} and \emph{negative} transfer. For the positive transfer
case the subtasks are the same---catching either ball results in a
reward of +1. In the negative transfer case the grey ball still gives a
reward of +1, but catching the white ball results in a reward of -1. In
this subtask the optimal agent must learn to catch the grey balls
\emph{and} avoid the perceptually-similar white balls; suboptimal
solutions would include avoiding or catching both types of balls. In
both setups the type of ball used is switched every episode.

Our baseline is the standard DQN. In order to provide a fair comparison,
we impose one condition on the architecture of our policy network, and
one condition on its training. For the first condition we divide the
number of neurons in the hidden layer of each option head by the number
of option heads, thereby keeping the number of parameters the same. For
the second condition we alternate heads when performing the Q-learning
update, keeping the number of training samples the same. We also
construct a ``half DQN'', which contains half the parameters of the
standard DQN in the fully connected hidden layer; this uses the standard
architecture, not the option heads. This tests whether the sample
complexity of our DQN with option heads is either the result of having
fewer parameters to tune in each head, or the result of our imposed
structural constraint. More details on the model architectures and
training hyperparameters are given in the Appendix.

As well as investigating the effects of different kinds of transfer, we
also look into the effects of varying the capacity of the
network---specifically we run experiments with 16, 32, and 64 neurons in
the fully connected hidden layer of the standard DQN. Correspodingly,
the half DQN has 8, 16, and 32 neurons, and each of the 2 option heads
in our network also has 8, 16, and 32 neurons.

\begin{figure*}[!ht]
  \includegraphics[trim={3cm 1.8cm 3cm 1.8cm}, clip, width=\textwidth,height=10cm]{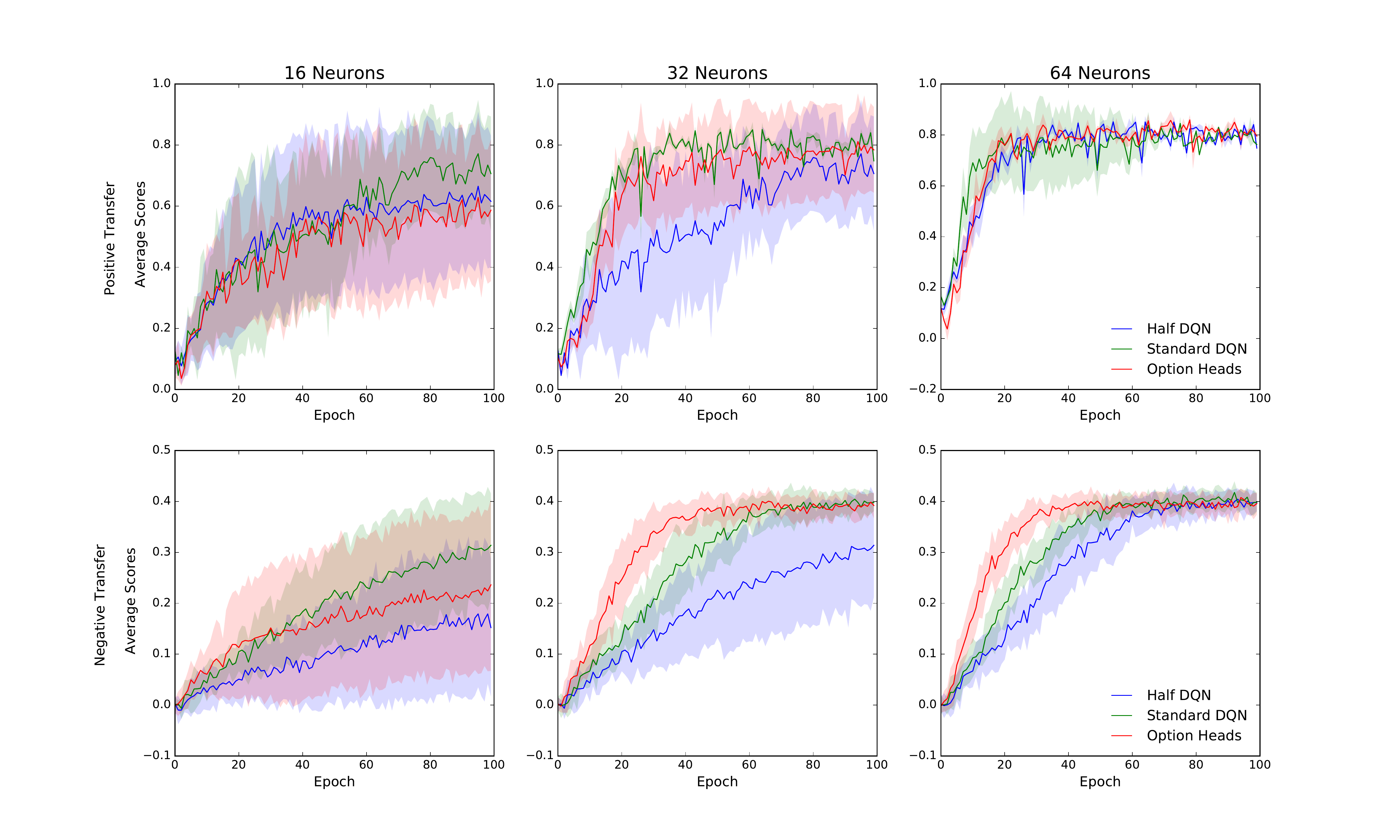}
  \caption{The average score per episode of the standard DQN, half DQN and DQN with 2 option heads. The average and standard deviation are calculated over 15 runs (with different random seeds), with the score for each run being averaged over 250 episodes. Each epoch corresponds to 10000 training steps. The number of neurons corresponds to the number of neurons in the fully connected hidden layer of the standard DQN. In the positive transfer setting the most important factor is capacity, not the architecture. As capacity increases the difference in performance between the networks diminishes. In the negative transfer setting the effect of capacity is strong when capacity is very low, but otherwise the DQN with option heads demonstrates a superior sample complexity over the baselines. Best viewed in colour.}
  \label{scores}
\end{figure*}

As seen in Figure \ref{scores}, lack of capacity has the most
significant effect on performance. As capacity increases, the
differences between the three networks diminishes. Besides capacity,
architecture does not appear to have any significant impact on the
positive transfer subtasks. However, with negative transfer subtasks the
DQN with option heads is able to make significantly quicker progress
than the standard DQN. When given enough capacity, our control for
``head capacity'' in this experiment---the half DQN---also converges to
the same policy in terms of performance, but with a larger sample
complexity. This suggests that incorporating domain knowledge in the
form of structural constraints can be beneficial, even whilst keeping
model capacity the same---in particular, the quicker learning suggests
that this knowledge can effectively be utilised to reduce the number of
samples needed in deep reinforcement learning.

Qualitatively, the convolutional filters learned by all DQNs are highly
similar. This reinforces the intuition that the structural constraint
imposed upon the DQN with option heads allows low-level feature
knowledge about falling balls and moving the paddle to be learned in the
shared convolution layers, whilst policies for catching and avoiding
balls are represented more explicitly in each head.

As the classification task for the supervisory network is simple in this
domain, we do not attempt to replace it with an oracle during
evaluation. In practice the network learns to divide the state space
rapidly.

\section{Discussion}\label{discussion}

We show that with a simple architectural adjustment, it is possible to
successfully impose prior domain knowledge about subtasks into the DQN
algorithm. We demonstrate this idea in a game of catch, where the task
of catching or avoiding falling balls depending on their colour can be
decomposed intuitively into the subtask of catching grey balls and
another subtask of avoiding white balls---subtasks that incur negative
transfer. We show that learning the subtasks separately on different
option heads allows the DQN to learn with lower sample complexity. The
shared convolutional layers learn generally useful features, whilst the
heads learn to specialise. In comparison, the standard DQN presumably
suffers from subtask interference with only a single Q-function.
Additionally, the structural constraint does not hinder performance when
learning subtasks with positive transfer.

Our results are contrary to those reported with the MCDQN trained on
eight Atari games simultaneously, which was outperformed by a standard
DQN \cite{parisotto2015actor}. According to Parisotto \emph{et al.}
\shortcite{parisotto2015actor}, the standard DQN and MCDQN tend to
focus on performing well on a subset of games at the expense of the
others. We posit that this ``strategy'' works well for the DQN, whilst
the explicitly constructed controller heads on the MCDQN receive more
consistent training signals, which may cause parameter gradients that
interfere with each other in the shared convolutional layers
\cite{caruana2012dozen,rusu2015policy}. This is less of a concern in
the hierarchical reinforcement learning setting as we assume a more
coherent state space across all subtasks.

In the regime of subtasks rather than multiple tasks, the DQN with
option heads is not as generalisable as goal-based approaches
\cite{schaul2015universal,kulkarni2016hierarchical}. However, these
methods currently require a hand-crafted representation of the goals as
input for their networks---a sensible approach only when the goal
representation can be reasonably appended to the state space. In
contrast, our oracle mapping from states to options can be used when
constructing representations of goals as inputs is not a straightforward
task. This suggests future work in removing the oracle by focusing on
\emph{option discovery}, where the agent must learn options without the
use of a supervisory signal.

\section*{Acknowledgements}
The authors gratefully acknowledge a gift of equipment from NVIDIA
Corporation and the support of the EPSRC CDT in Neurotechnology.

\bibliographystyle{named}
\bibliography{../references.bib}

\appendix
\part*{Appendix}
\section{Model}\label{model}

We use a smaller DQN than the one specified in \cite{mnih2015human}.
The baseline (standard) DQN architecture that we used in our
experiments, with a capacity of ``32 neurons'', is as follows:

\begin{scriptsize}
\begin{center}
\begin{tabular}{c p{7.0cm}}
    \hline
    \textbf{Layer} & \textbf{Specification}\\
    \hline\hline
    1 & 32 5x5 spatial convolution, 2x2 stride, 1x1 zero-padding, ReLU \\ \hline
    2 & 32 5x5 spatial convolution, 2x2 stride, ReLU \\ \hline
    3 & 32 fully connected, ReLU \\ \hline
    4 & 3 fully connected \\
    \hline 
  \end{tabular}
\end{center}
\end{scriptsize}

As our version of Catch can be divided into 2 distinct subtasks, our
policy network therefore has 2 option heads. Each option head has 16
neurons each in the penultimate fully connected layers---half that of
the baseline DQN. The half DQN therefore also has 16 neurons in the
penultimate fully connected layer. The same formula is employed when
testing with a capacity of 16 and 64 neurons.

Unlike \cite{osband2016deep}, we do not normalise the gradients coming
through each option head, as the errors are only backpropagated through
one head at a time.

\section{Hyperparameters}\label{hyperparameters}

Hyperparameters were originally manually tuned for the model with the
original version of Catch \cite{mnih2014recurrent} (only white balls
giving a reward of +1). We then performed a hyperparameter search over
learning rates \(\in\) \{0.000125, 0.00025, 0.0005\}, target network
update frequencies \(\in\) \{4, 32, 128\}, and final values of
\(\epsilon\) \(\in\) \{0.01, 0.05\}. The learning rate with the best
performance for all models was 0.00025, except for the standard DQN and
half DQN in the negative transfer setting, where 0.000125 was better.
This is an interesting finding---with negative transfer the standard
architecture requires a lower learning rate, whilst with option heads a
higher learning rate can still be used. The following hyperparameters
were used for all models, where only hyperparameters that differ from
those used in \cite{van2015deep} for the tuned double DQN are given:

\newcommand{\specialcell}[2][c]{
  \begin{tabular}[#1]{@{}l@{}}#2\end{tabular}
}

\begin{scriptsize}
\begin{center}
  \begin{tabular}{p{2.3cm}p{0.7cm}p{4.5cm}}
    \hline
    \textbf{Hyperparameters}  & \textbf{Value} & \textbf{Description}                                                                                           \\ \hline\hline
    \specialcell[t]{Replay \\memory size}              & 10000 & Size of each experience replay memory buffer.                                                                           \\ \hline
    \specialcell[t]{Target network\\update frequency} & 4     & Frequency (in number of steps) with which the target network parameters are updated with the policy network parameters. \\ \hline
    Optimiser                       & Adam  & Stochastic gradient descent optimiser.                                                                                  \\ \hline
    \specialcell[t]{Final exploration \\frame}     & 10000 & Number of steps over which $\epsilon$ is linearly annealed.                                                             \\ \hline
    \specialcell[t]{Replay start \\size}               & 10000 & Number of steps of random exploration before $\epsilon$ is annealed.                                                    \\ \hline
    \specialcell[t]{Gradient \\clipping }              & 10    & Max absolute value of the L2 norm of the gradients.                                                                     \\ \hline
    \specialcell[t]{Validation \\frequency}            & 10000 & Number of steps after which evaluation is run.                                                                          \\ \hline
    \specialcell[t]{Validation \\steps }               & 6000  & Number of steps to use during evaluation. Corresponds to 250 episodes of Catch.                                         \\ \hline
  \end{tabular}
\end{center}
\end{scriptsize}

\end{document}